\documentclass[runningheads]{llncs}
\usepackage{graphicx}
\usepackage{subcaption}
\usepackage{xcolor}
\begin{document}
\title{Extracting Digital Biomarkers for Unobtrusive Stress State Screening from Multimodal Wearable Data}

\author{Berrenur Saylam\inst{1}\orcidID{0000-0002-8294-8342} \and Özlem Durmaz İncel\inst{1}\orcidID{0000-0002-6229-7343} }
\authorrunning{B. Saylam, Ö. Durmaz Incel}
% If there are more than two authors, 'et al.' is used.%
\institute{Boğaziçi University, İstanbul 34342, Turkey \\
\email{\{berrenur.saylam, ozlem.durmaz\}@boun.edu.tr}\\
%\url{http://www.springer.com/gp/computer-science/lncs} 
}
\maketitle              % typeset the header of the contribution
\begin{abstract}
With the development of wearable technologies, a new kind of healthcare data has become valuable as medical information. These data provide meaningful information regarding an individual's physiological and psychological states, such as activity level, mood, stress, and cognitive health. These biomarkers are named digital since they are collected from digital devices integrated with various sensors. In this study, we explore digital biomarkers related to stress modality by examining data collected from mobile phones and smartwatches. We utilize machine learning techniques on the Tesserae dataset, precisely Random Forest, to extract stress biomarkers. Using feature selection techniques, we utilize weather, activity, heart rate (HR), stress, sleep, and location (work-home) measurements from wearables to determine the most important stress-related biomarkers. We believe we contribute to interpreting stress biomarkers with a high range of features from different devices. In addition, we classify the $5$ different stress levels with the most important features, and our results show that we can achieve $85\%$ overall class accuracy by adjusting class imbalance and adding extra features related to personality characteristics. We perform similar and even better results in recognizing stress states with digital biomarkers in a daily-life scenario targeting a higher number of classes compared to the related studies. 

\end{abstract}
\section{Introduction}
Clinical biomarkers are objectively-measured and evaluated indicators of the biological processes~\cite{a1,a2}. The biomarkers and the clinical outcomes aim to assist in understanding the factors affecting human health.  

It becomes possible to measure health parameters, such as the amount of daily activity, sleep quality, and stress level while working on a task with wearable devices, such as smartwatches, sleep sensors, and step counters, in an ambulatory and continuous manner. This digitalization of the healthcare measurements created the term \textit{digital biomarkers}, which are objective, quantifiable, physiological, and behavioral measures collected from wearables and smartphones~\cite{a3}. The indicators or characteristics for a digital marker could be computed from one or more digital health technologies. 

 Furthermore, there are different categories of biomarkers in literature~\cite{a3}, such as \textit{diagnostic biomarkers} to detect any disease or condition~\cite{a4}, \textit{pharmacodynamic response biomarkers} to measure changes in response to pharmaco agents~\cite{a5} and \textit{monitoring biomarkers} to monitor a specific medical condition~\cite{a6}. Thus, there are mainly three types of physiological outcomes. Outcomes for all three can be measured via sensors. Additionally, outcomes can be measured via questionnaires for the second type, i.e., \textit{pharmacodynamic response biomarkers}. Our study falls into the monitoring biomarkers category since data is collected during participants' daily lives.

In this work, we examine Tesserae dataset~\cite{a13}, which is collected from office workers using smartphones for activity tracking, wearable devices for step-count, sleep and heart rate measurement, and Bluetooth beacons to understand locations such as home, work, and questionnaires for ground truth information. 
Previous studies which are using the Tesserae dataset (Table~\ref{tab:Tesserae}) focus on understanding the effects of seasons and weather on sleep patterns~\cite{a14}, predicting job performance~\cite{a15}, and predicting mental health from social network data~\cite{a16}. However, this work focuses on stress level classification and examines the affecting biomarkers. To the best of our knowledge, there is no work considering these aspects on this dataset.

We investigate the extraction of novel biomarkers related to stress with various parameters and modalities, such as mobile phones and wearable devices. We aim to predict the participants' stress levels using the data from wearables and smartphones and find the most important biomarkers, in other words, the features related to stress levels. We rank the features to show which modalities, such as sleep, and activity levels, are more important in classifying stress levels. We have daily stress responses from the participants as the labels. Using these labels, we classified the five stress levels from low to high provided in ground truth data with these important parameters and compared the result with the usage of all parameters using the Random Forest (RF) algorithm. We also investigate the inclusion of parameters related to the participants' personalities. 

Since the number of examples in each class is not the same, and some classes have fewer instances, we observed lower recognition performance for these classes. We applied SMOTE (synthetic minority over-sampling) to solve the class imbalance problem. We obtained the best classification result, $85\%$ overall class accuracy, by adding the personality-related parameters in our parameter space and solving the class imbalance issue. In the end, in all cases, we found that the most important parameters are sleep and phone activity modalities. As far as we examined, this is the first study extracting digital biomarkers, in other words, features related to stress state from the Tesserae dataset.

The rest of the paper is organized as follows. In Section \ref{sec:Related}, we explore literature related to biomarkers and stress monitoring. In Section \ref{sec:Method}, we explain the details of the dataset, the data construction step for our analysis, the data preprocessing step before analysis, and details of the target class, i.e., stress. In Section \ref{sec:results}, we provide obtained results with feature ranking, modality ranking, and performance details with a discussion and comparison to the related studies. In Section \ref{sec:Conclusion}, we explain our findings and discussion points for further analysis.

\begin{table}[!b]
%\addtolength{\tabcolsep}{-1pt}
\centering
\caption{Studies using the Tesserae dataset in the literature}
\label{tab:Tesserae}
\centering
\resizebox{\textwidth}{!}{
\begin{tabular}{|c|c|c|c|}
\hline
\textbf{Study}              & \textbf{Aim}                                                                                                                        & \textbf{Methodology}                                                                                                                                                                                                                                                                                                                                                       & \textbf{Result}                                                                                                                                                                              \\ \hline
\cite{a14}                 & \begin{tabular}[c]{@{}c@{}}Investigation of the effects of\\ seasons and weather on sleep\end{tabular}                              & \begin{tabular}[c]{@{}c@{}}Construction of an independent model\\ for each variable which is sleep duration, \\ bedtime, and wake-up time.\\ Usage of mixed linear effect model.\end{tabular}                                                                                                                                                                              & \begin{tabular}[c]{@{}c@{}}The strongest effect on wake time \\ and sleep duration\\ especially in the spring season\end{tabular}                                                            \\ \hline
\cite{a15}                 & \begin{tabular}[c]{@{}c@{}}Prediction of workers job \\ performance\end{tabular}                                                    & \begin{tabular}[c]{@{}c@{}}Extraction of the high-level features with\\ AutoEncoder (AE) usage and gradient\\ analysis to understand the causality\\ relation between features and the target\end{tabular}                                                                                                                                                                 & \begin{tabular}[c]{@{}c@{}}Prediction of job performance\\ with \%75 f1-score.\\ Analyze of different job \\ performance metrics parameters' effects\\ on the model performance\end{tabular} \\ \hline
\cite{a16} & \begin{tabular}[c]{@{}c@{}}Prediction of mental health,\\ depression and anxiety,\\ throughout the social network data\end{tabular} & \begin{tabular}[c]{@{}c@{}}Application of network analysis methods\\ for different types of tasks \\ such as depressed and non-depressed\\ positions in network differ or not, \\ difference between network positions\\ lead to depression or anxiety trait difference,\\ dynamic or static network lead to better\\ results on the mental health prediction\end{tabular} & \begin{tabular}[c]{@{}c@{}}Found that the inclusion of dynamics\\ leads to further improvements to the\\  static network data analysis results\end{tabular}                                                \\ \hline
\end{tabular}
}
\end{table}

\section{Related Works}
\label{sec:Related}

Many related works exist in the literature about biomarkers for various physiological and psychological states~\cite{a7,a8,a9,a10,a11,a12}.
In~\cite{a7}, the authors examined the behavioral markers to track physical activity following hospital discharge via data from wearable devices. The idea is to understand the characteristics of the individuals in smartphone and wearable device usage to monitor health-related measurements. They identified four biomarker sets i) more agreeable and conscientious; ii) more active, social, and motivated; iii) more risk-taking and less supported; and iv) less active, social, and risk-taking. Identification of these groups of biomarkers has been made using latent class analysis.

In~\cite{a8}, researchers examined biomarkers of depression state outside of the laboratory. Similar to~\cite{a7}, they collected data with wearable devices and state-of-the-art questionnaires, namely Patient Health Questionnaire (PHQ-9), as a ground truth. PHQ-9 is a standard self-report for depression assessment in clinical contexts~\cite{a9}. As a methodology, they used multiple regression analysis to predict the PHQ-9 result and find the parameter which can significantly predict the target value, in this case, the PHQ-9 value.  

In another study~\cite{a10}, the main idea was to extract the underlying effects of seasons and weather parameters on the sleep data. Similar to the previous work~\cite{a8}, they utilized regression-based modeling. They focused on continuous data collection via wearable devices to differentiate their study from state-of-the-art.

Even though statistical approaches for extracting biomarkers are common in the recent literature, some studies use simple machine-learning (ML) techniques~\cite{a11,a12}. For instance, in~\cite{a11}, the aim was to determine digital biomarkers for frailty, an essential factor in older adults' recovery process. They examined physical activity parameters, such as percentage time standing, percentage time walking, walking cadence, and longest walking bout, to identify slowness, weakness, exhaustion, and inactivity classes of physical frailty. As a methodology, they utilized ANOVA and binary logistic regression model. 

As another example of the usage of ML models in extracting biomarkers, a study~\cite{a12} on stress and mental health status markers extraction can be given. Again, wearable device measurements are analyzed to determine the self-reported values—this study, different from others, utilized SVM models with different configurations. 

We observe that recent studies on biomarkers are utilizing either basic statistical methods, such as ANOVA analysis, or different forms of regression models. When utilizing ML models, state-of-the-art also uses traditional models. 

In addition to the biomarkers' literature, there are many studies on stress detection and monitoring~\cite{b1,b2,b3,a17}.

In~\cite{b1}, authors examined features from multi-modal smartphone sensors, i.e., accelerometer sensor for physical activity, microphone for social interaction, phone calls and application usage for social activity, etc. They proposed combinations of machine learning techniques such as semi-supervised learning, ensemble methods, and transfer learning to classify three stress levels (low, mid, high).

In~\cite{b2}, researchers added surrounding data to personal stress data to examine whether these improve overall stress detection accuracy. Again, they used three-level stress for classification. They used low-level features from smartphone sensor data, such as average, standard deviation, minimum, maximum, etc. They designed different scenarios on their data and StudentLife~\cite{b3} data alone and in combination with personal and surroundings data. They reached the highest performance when they combined personal and surrounding data, $79.16\%$ and $81.79\%$ on their collection of data and StudentLife, respectively.

Moreover, stress-related studies also can be classified according to the data collection environment, such as in a controlled laboratory and daily (unobtrusively) environment. In~\cite{a17}, authors extensively explain state-of-the-art using different signal types in different environments. Depending on the modality, different sensor types for measuring physiological changes exist. For instance, accelerometer, body temperature, EDA (electrodermal activity), HR (heart rate), HRV (heart rate variability) \cite{b18}, and speech are some of the typical sensor types for physiological changes. There is also usage of high-level features extracted from mobile phone usage patterns, physical activity (total number of steps), and screen events.

We observe that literature studies validate their proposed selection of sensor types, data collection environment, and proposed algorithm type. In this study, we contribute to the parameter space of the stress features by examining an extensive set of multi-modal data analyses.

\newpage
\section{Methodology}
\label{sec:Method}
\subsection{Dataset}
In this study, we utilize Tesserae Dataset\footnote{https://tesserae.nd.edu/}~\cite{a13}. It is collected to measure office workers' workplace performance via psychological traits and physical characteristics over one year. It consists of data from 757 subjects. A smartphone and Garmin watch are used as data collecting devices. 

Data were collected from the phones with an activity tracking application. The participants wore a Garmin watch during the study period. It has 5-7 days of battery life. In addition, Bluetooth beacons have been used to extract information about the location, such as home or office. As complementary data to the non-verbal measurements by devices, they also collected verbal data from social media such as Facebook and LinkedIn, which were not accessible to us. 

The University of Notre Dame approved this data collection campaign as a research project, and a consent form has been taken from the participants.

\subsection{Ground Truth}
For collecting ground truth data, various questionnaires were shared with the participants. These questionnaires are related to job performance, intelligence, mood, anxiety, health measure, exercise, sleep, and stress. At the beginning of the study, participants filled out all questions for each type of survey. The required duration is about one hour to fill. In addition, there are daily survey questions that constitute the summary of each type of measurement. Each questionnaire's overall score corresponds to one column in the daily survey scores file. Thus, it requires only a few minutes to fill every day. Moreover, they collected an end study questionnaire and follow-up survey. However, they were not accessible to us. Thus, the questionnaires have four main parts as in the following.
\begin{itemize}
    \item Initial Ground Truth
    \item Daily Surveys
    \item Exit 
    \item Follow-up Survey
\end{itemize}

To give an idea about the content of the questionnaires, we summarize them in Table \ref{table:survey}. We show the ones we have access to using * mark.

\begin{table}[!htb]
%\addtolength{\tabcolsep}{-5.2pt}
\centering
\caption{Details of the questionnaires}
\begin{tabular}{llllll}
\hline
\multicolumn{1}{|l|}{\textbf{Questionnaires}}                                                         & \multicolumn{5}{c|}{\textbf{Content}}                                                                                                                                                                                                                                                                                                                                                                                                                                                                                                                                                                                   \\ \hline
\multicolumn{1}{|l|}{\textit{\begin{tabular}[c]{@{}l@{}}Initial Ground \\ Truth (IGTB)\end{tabular}}} & \multicolumn{5}{c|}{\begin{tabular}[c]{@{}c@{}}In-Role Behavior (IRB), Individual Task Proficiency (ITP), \\ Organizational Citizenship Behavior (OCB), \\ Interpersonal and Organizational Deviance Scale (IOD), \\ Shipley Abstraction*,  Shipley Vocabulary*, \\ Big Five Inventory* (BFI), \\ Positive and Negative Affect* Schedule (PANAS), \\ State-Trait Anxiety Inventory* (STAI), \\ MITRE modified Alcohol Use Disorders (AUDIT MITRE), \\ Modified Global Adult Tobacco Survey (Modified GATS),  \\ International Physical Activity Questionnaire* (IPAQ), \\ Pittsburgh Sleep Quality Index* (PSQI)\end{tabular}} \\ \hline
\multicolumn{1}{|l|}{\textit{Daily Surveys}}                                                          & \multicolumn{5}{c|}{\begin{tabular}[c]{@{}c@{}}In-Role Behavior (IRB), Individual Task Proficiency (ITP), \\ Organizational Citizenship Behavior (OCB),\\ Counterproductive Workplace Behavior (CWB), \\ Big Five Inventory* (BFI), \\ Positive and Negative Affect Schedule* (PANAS), \\ Omnibus Anxiety Question*, Tobacco Use Assessment, \\ MITRE Omnibus Stress Question*, \\ MITRE Alcohol Use Assessment, \\ MITRE Physical Activity Assessment*\\ MITRE Sleep Assessment*, \\ MITRE Context Assessment\end{tabular}}                                                                                                  \\ \hline
\multicolumn{1}{|l|}{\textit{Exit Battery}}                                                           & \multicolumn{5}{c|}{\begin{tabular}[c]{@{}c@{}}State-Trait Anxiety Inventory (STAI), \\ Positive and Negative Affect Schedule (PANAS), \\ Morningness-Eveningness Questionnaire (MEQ), \\ Emotional Regulation, Emotional Intelligence, \\ Pro-social Motivational\end{tabular}}                                                                                                                                                                                                                                                                                                                                        \\ \hline
\multicolumn{1}{|l|}{\textit{Followup}}                                                               & \multicolumn{5}{c|}{Life events, sick days, vacation days}                                                                                                                                                                                                                                                                                                                                                                                                                                                                                                                                                              \\ \hline
                                                                                                      &                                                                                                                        &                                                                                                                        &                                                                                                                       &                                                                                                                       &                                                                                                                       \\
                                                                                                      &                                                                                                                        &                                                                                                                        &                                                                                                                       &                                                                                                                       &                                                                                                                       \\
                                                                                                      &                                                                                                                        &                                                                                                                        &                                                                                                                       &                                                                                                                       &                                                                                                                       \\
                                                                                                      &                                                                                                                        &                                                                                                                        &                                                                                                                       &                                                                                                                       &                                                                                                                       \\
                                                                                                      &                                                                                                                        &                                                                                                                        &                                                                                                                       &                                                                                                                       &                                                                                                                       \\
                                                                                                      &                                                                                                                        &                                                                                                                        &                                                                                                                       &                                                                                                                       &                                                                                                                      
\end{tabular}
\label{table:survey}
\end{table}

\subsection{Dataset Construction}
In the shared data file, we did not have access to all measurements as stated in the previous section. We made our analysis on the reduced data. Even if detailed sampling measurements are sampled from the wearable device from 3 minutes to 15 minutes, depending on the data modalities, we made our analysis on daily bases. The used file details can be seen in Table \ref{table:ConstructedAnalysis}.

\begin{table}[!t]
\centering
%\addtolength{\tabcolsep}{-3.9pt}
\caption{Details of the constructed data for analysis}
\begin{tabular}{|l|lllll|}
\hline
\textbf{Data Folder}                                              & \multicolumn{5}{c|}{\textbf{Used Data File}}                                                                                                                                                                                                             \\ \hline
\textit{Context}                                                  & \multicolumn{1}{l|}{Weather (78)}                                                    & \multicolumn{1}{l|}{}                                                               & \multicolumn{1}{l|}{}        & \multicolumn{1}{l|}{}            &           \\ \hline
\textit{\begin{tabular}[c]{@{}l@{}}Garmin\\ Summary\end{tabular}} & \multicolumn{1}{l|}{Activity (8)}                                                    & \multicolumn{1}{l|}{Daily (25)}                                                     & \multicolumn{1}{l|}{HR (11)} & \multicolumn{1}{l|}{Stress (11)} & Sleep (7) \\ \hline
\textit{Mixed}                                                    & \multicolumn{1}{l|}{\begin{tabular}[c]{@{}l@{}}Phone \\ Activity (128)\end{tabular}} & \multicolumn{1}{l|}{\begin{tabular}[c]{@{}l@{}}HR WorkDesk\\ Home (7)\end{tabular}} & \multicolumn{1}{l|}{}        & \multicolumn{1}{l|}{}            &           \\ \hline
\textit{\begin{tabular}[c]{@{}l@{}}Ground \\ Truth\end{tabular}}  & \multicolumn{1}{l|}{\begin{tabular}[c]{@{}l@{}}Daily \\ Scores (19)\end{tabular}}    & \multicolumn{1}{l|}{}                                                               & \multicolumn{1}{l|}{}        & \multicolumn{1}{l|}{}            &           \\ \hline
\end{tabular}
\label{table:ConstructedAnalysis}
\end{table}

Even though they collected data from $757$ participants, we had access to $727$ participants' data. Furthermore, although it is stated\footnote{https://osf.io/yvw2f/wiki/EMAs/} that $56$ days of the daily survey is collected, we observed $61$ days of daily survey answers from some of the participants. To construct our dataset, we merged the files stated in Table \ref{table:ConstructedAnalysis} according to \textit{ParticipantID} and \textit{Timestamp}. We obtained wearable and ground truth data for each participant on a daily basis. While merging, we noticed that for some participants, the amount of data from wearables is more than ground truth survey answers (it means they collected more data with the devices than the planned data collection period). However, as we need the ground truth information for our analysis, we considered participants' data that only have ground truth responses, i.e., responses to the daily surveys. After constructing the data file, we have 36294 rows of data coming from all participants. If we had $61$ days of ground truth for everyone, it must be $44347$ rows of data. Thus, it is clear that we do not have survey answers from all participants for precisely $61$ days. In addition, we merged files by considering the common columns in each file. Thus, as we have common columns in each file, such as \textit{ParticipantID}, \textit{Timestamp}, and \textit{Date}, we did not add them in the final construction since a one-time indication is sufficient in the whole dataset. For this reason, we obtained fewer columns ($269$) compared to the sum of each data file's column number ($294$), which are given in detail in Table \ref{table:ConstructedAnalysis} between the parentheses. In the final data, we have $269$ columns where $15$ of them are the ground truth. Here, $15$ columns coming from the ground truth which are \textit{survey name}, \textit{stress}, \textit{anxiety}, \textit{sleep}, \textit{positive affect}, \textit{negative affect}, \textit{extraversion}, \textit{agreeableness}, \textit{conscientiousness}, \textit{neuroticism}, \textit{openness}, \textit{total phone activity duration}, \textit{survey sent time}, \textit{survey start time}, \textit{survey finish time}. The rest of the columns from the other modalities are not given explicitly since their volume is quite large. In this section, we explained only the construction of the data. We preprocess it for further analysis, which is explained in Section \ref{sec:Preprocess}.

\subsection{Details of the stress classes}
\begin{figure}[!htbp]
\centerline{\includegraphics[width=0.78\columnwidth]{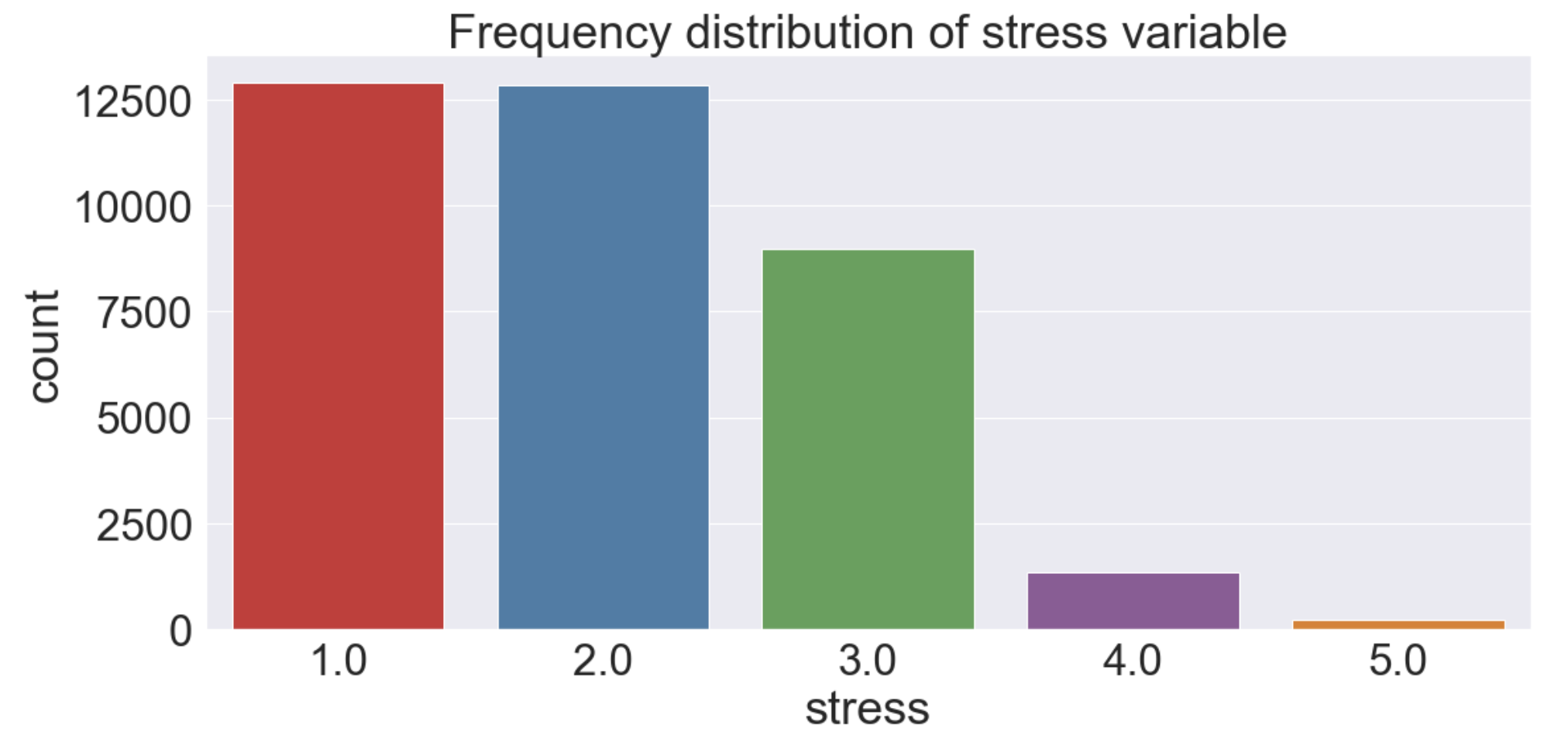}}% 58 onlyStressFreq
\caption{Distribution of stress classes}
\label{stressFreq}
\end{figure}

We selected stress as our dependent variable in the scope of this study, and we examined its distribution in detail. In Figure \ref{stressFreq}, one can see that there is a class imbalance. We have a few instances from some classes, especially from $4th$ and $5th$ classes. The model may fail to recognize them due to insufficient examples. To solve this issue, we applied SMOTE technique. After the implementation, we got $57150$ rows for both our experiments which are explained in detail in Section \ref{sec:Preprocess}, and column numbers remained the same. Here, classes 1 and 5 correspond to very low stress and very high-stress levels, respectively. To show the effect on the model performance, we will provide model results with class imbalance along with performance after applying SMOTE in Section \ref{sec:results}.

\subsection{Data Preprocessing}
\label{sec:Preprocess}
We deleted irrelevant attributes for our analysis, such as \textit{survey name}, \textit{survey sent time}, \textit{survey start time}, and \textit{survey end time}. We selected the stress variable among ground truth variables as a target variable. Nine missing values were in the stress column over all participants' data, and we deleted rows corresponding to them rather than applying an imputation method for simplicity. 

We also examined missing values among the independent variables. Some attributes were never collected from some participants. Thus, the imputation methodology could not be applied for these attributes. Among them, if there is a lack of many participants' data, their removal by row leads to a high decrease in our dataset. Instead, we deleted those columns. These are \textit{act (activity) still}, \textit{light mean}, \textit{garmin hr (heart rate) min}, \textit{garmin hr max}, \textit{garmin hr median}, \textit{garmin hr mean}, \textit{garmin hr std}, \textit{ave (average) hr at work}, \textit{ave hr at desk}, \textit{ave hr at desk}, \textit{ave hr not at work}, \textit{call in num (number)}, \textit{call in duration}, \textit{call out num}, \textit{call out duration}, \textit{call miss num}, and their derivatives according to time episodes. However, when we have data columns with missing values at the person level, we imputed them by applying the mean operation for each participant. After this preprocessing step, we got $31772$ rows and $195$ columns. 

In order to see the impact of personality attributes in our prediction analysis, we designed two experiments. In the first experiment, the ground truth attributes other than the stress level; extraversion, agreeableness, conscientiousness, neuroticism, openness, anxiety, sleep, positive affect, and negative affect were deleted from the constructed data file. Since we did not include personality-related columns from big five personality survey, we named this dataset as \textit{without personality dataset}. Thus, we have $31772$ rows and $185$ columns. In the second experiment, we included these parameters in our feature space for the analysis rather than removing them as in the first experiment. Since we have personality information in the constructed data file, it is named as \textit{with personality dataset}. Please note that as additional information, it does not only include personality-related attributes, which are extraversion, agreeableness, conscientiousness, neuroticism, and openness; there are also anxiety, sleep, positive affect, and negative affect attributes. In the end, we have $31772$ rows and $195$ columns. The last column is the stress level, our target variable, and the remaining $194$ columns include the features.

\subsection{Classifier and Validation}
As a classification method, we used Random Forest (RF) algorithm in our analyses because it is an ensemble method and perform better among the other used methods in literature in this domain \cite{a17}. We applied hyper-parameter optimization with quite a large space and found that in scikit-learn n\_estimators $1000$, criterion $gini$, min\_sample\_split $2$, min\_samples\_leaf $1$, max\_features $sqrt$ combination reveal the best performance. Thus, we continued with that setup. In addition, we ran RF with different train/test split sizes. As we increased train size, we got higher accuracy scores. However, to overcome over-fitting, $80\%$ and $20\%$ train and test size are chosen, respectively.

\section{Results}
\label{sec:results}

\subsection{Results on Without Personality dataset }
In this section, we work on data with $185$ columns where there are no personality information-related columns and other attributes such as anxiety, sleep, and positive and negative affect. First, we present the results of the imbalanced stress class. We provide the most important feature ranking, i.e., biomarkers and corresponding classification results. Then, we list important features and classification results when we solve imbalance via SMOTE application. 
\subsubsection{With class imbalance}

\begin{itemize}
    \item \textbf{Most important biomarkers}\\
     We started with $184$ independent biomarkers. We want to select only the important parameters within the parameter space that greatly impact the target, i.e., stress. Therefore, we applied Random Forest feature selection to extract the most important biomarkers. The most important $20$ features are shown in Figure \ref{featuresWithClassImbalance}. To see the effect of modalities, we rank these biomarkers according to their corresponding modalities by assigning an importance value, e.g., $20$ for the most important one and subtracting by one for the up-comings. Then, we sum and obtain modality ranking in Figure \ref{fig:sub-firstWithoutPersonality}. There, we see that the most important features come from phone activity, sleep, HR, stress, activity, and daily modalities calculated according to Table \ref{table:ConstructedAnalysis}.

    In Figure \ref{fig:ComparisonWithoutPersonalitySMOTE}, we see significant changes between stress classes for each type of marker. The amount of the effects are different and can be seen in Figure \ref{featuresWithClassImbalance}. For instance, the most important marker is wake-up time according to the feature ranking. When we examine its relation with stress classes in Figure \ref{fig:WakeupStressWithoutPersonality}, we notice that the wake-up time interval becomes narrow concerning the increase in the stress level. Thus, it is found that the feature ranking affects the stress level. %and expected to observe it in Figure \ref{featuresWithClassImbalance}.
    Please note that \textit{stress} feature, from Garmin based on heart rate variability measures (the amount of time between consecutive heartbeats), has an important effect on the target stress class. It can be interpreted that the stress measurements from the Garmin confirm the subjective stress score collected as ground truth.

        \begin{figure}[!htbp]
        \centerline{\includegraphics[width=\columnwidth]{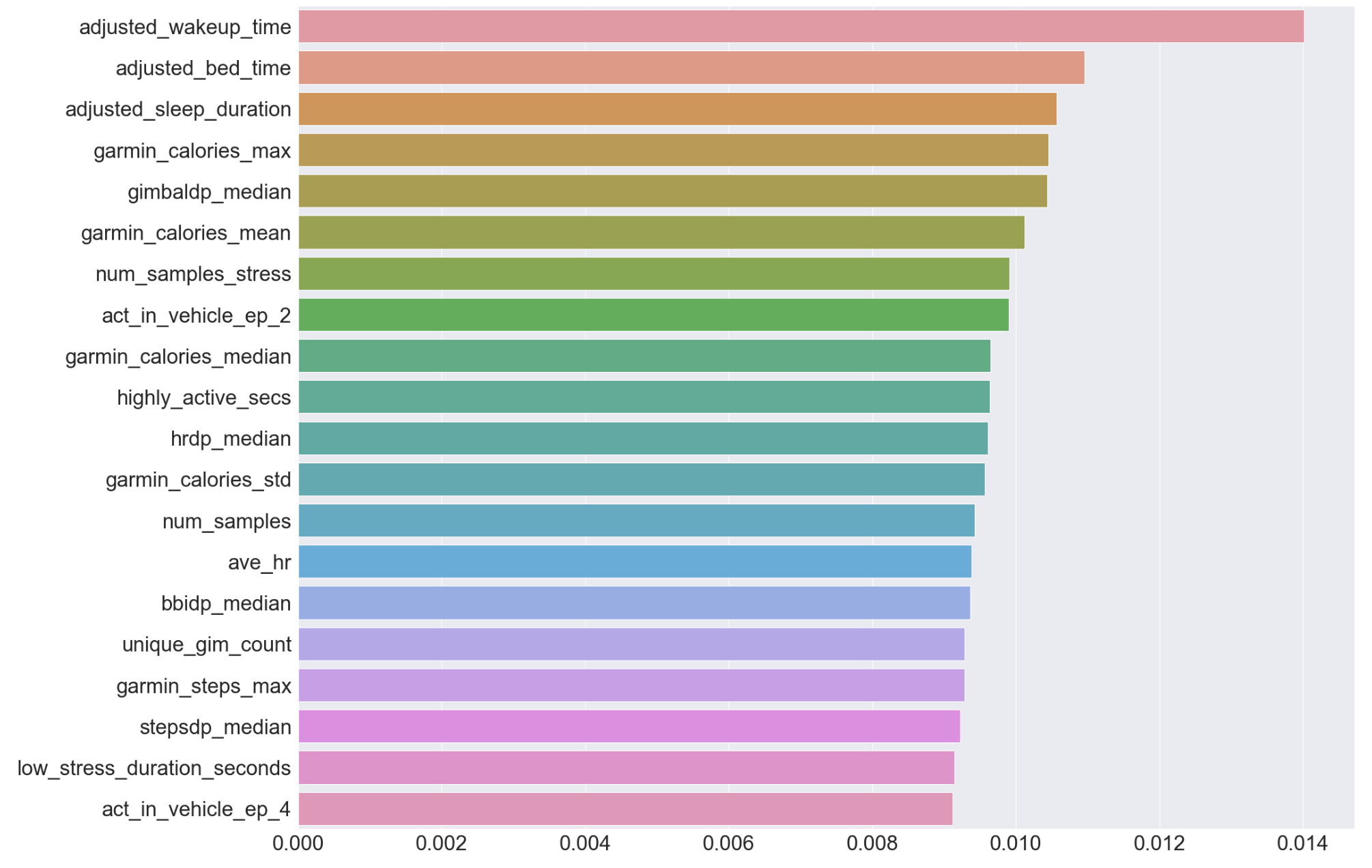}}
        \caption{Feature ranking with class imbalance without personality}
        \label{featuresWithClassImbalance}
        \end{figure}

    \begin{figure}[!htb]
\begin{subfigure}{.5\textwidth}
  \centering
  % include first image
  \includegraphics[width=\linewidth]{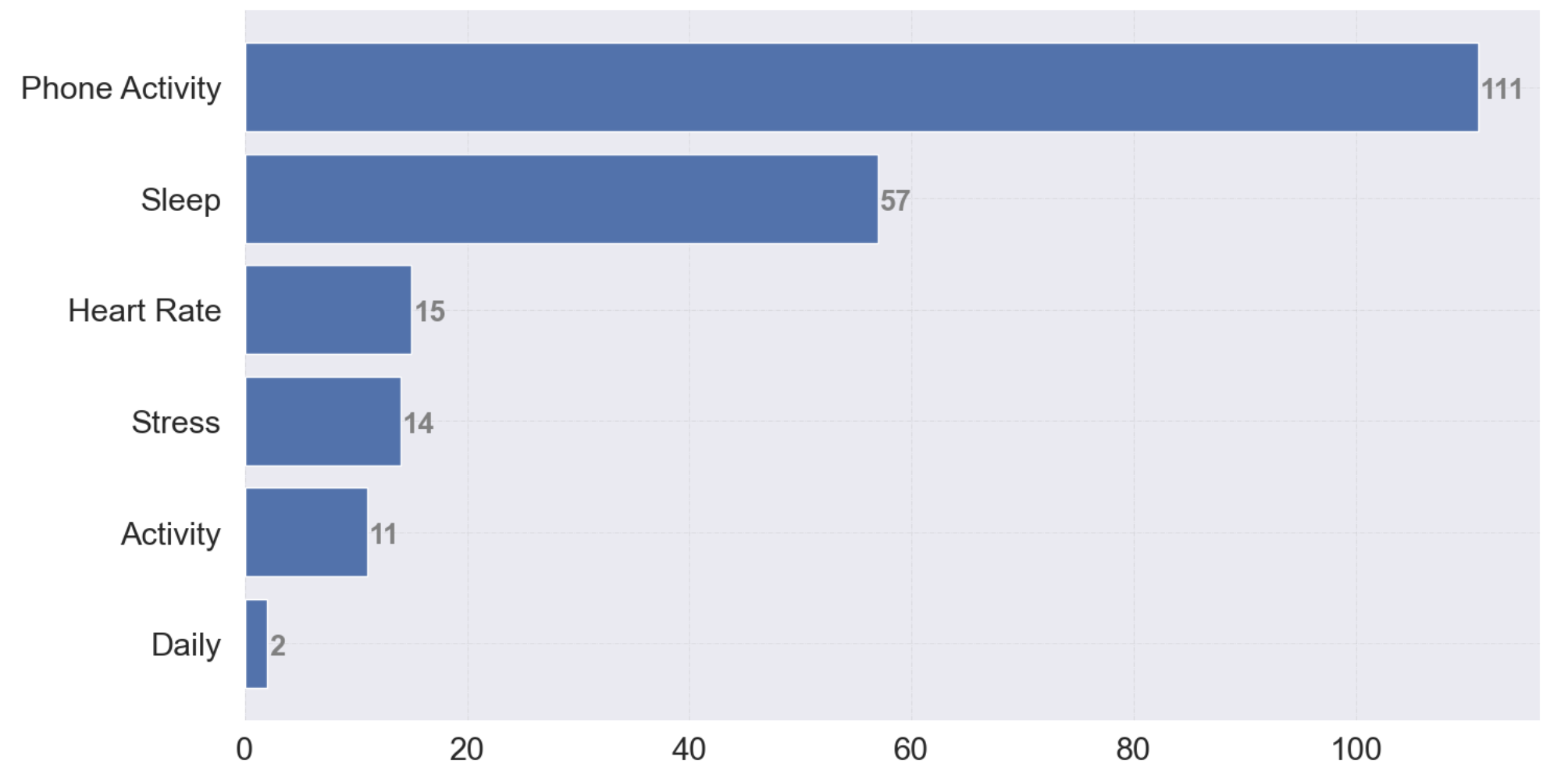}  
  \caption{With Class Imbalance}
  \label{fig:sub-firstWithoutPersonality}
\end{subfigure}
\begin{subfigure}{.5\textwidth}
  \centering
  % include second image
  \includegraphics[width=\linewidth]{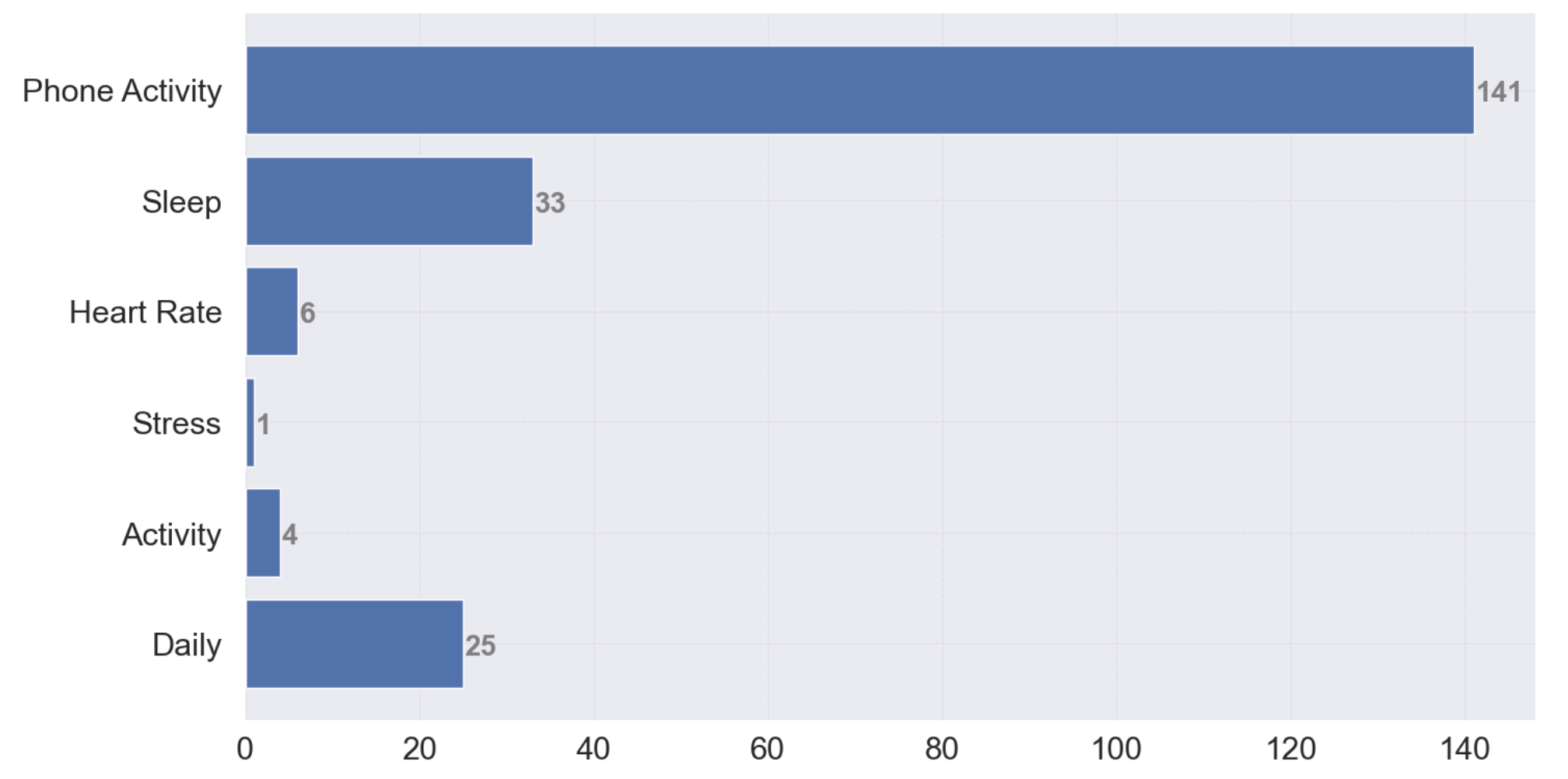}  
  \caption{After SMOTE application}
  \label{fig:sub-secondWithoutPersonality}
\end{subfigure}
\caption{Ranking of biomarkers' corresponding modalities: Without Personality Dataset}
\label{fig:RankingBiomarkerModalitiesWithoutPersonality}
\end{figure}

\begin{figure}
\begin{subfigure}{.5\textwidth}
  \centering
  % include first image
  \includegraphics[width=.8\linewidth]{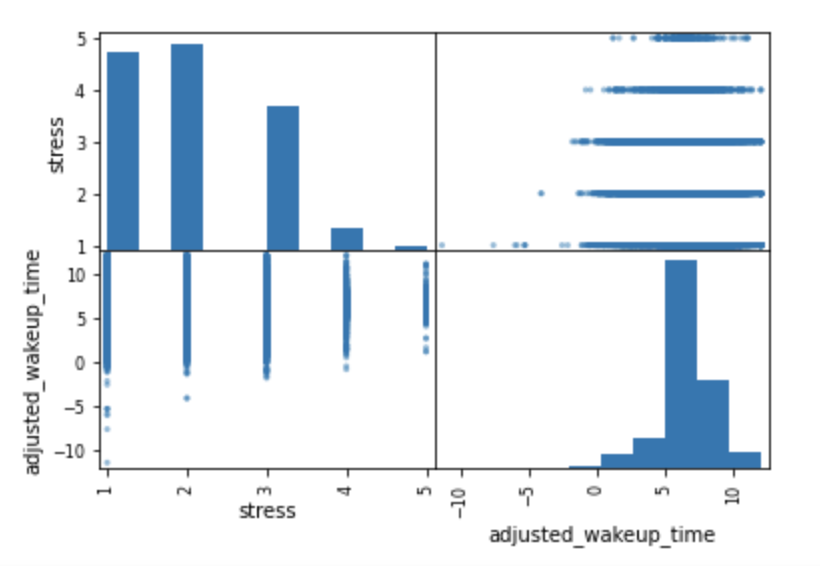}  
  \caption{Wake-up time vs Stress}
  \label{fig:WakeupStressWithoutPersonality}
\end{subfigure}
\begin{subfigure}{.5\textwidth}
  \centering
  % include second image
  \includegraphics[width=.8\linewidth]{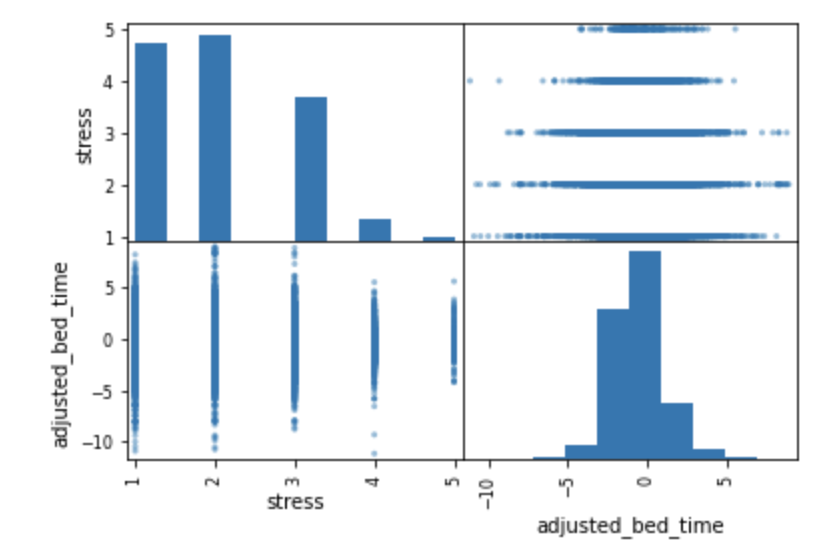}  
  \caption{Bed time vs Stress}
  \label{fig:sub-second}
\end{subfigure}

\begin{subfigure}{.5\textwidth}
  \centering
  % include third image
  \includegraphics[width=.8\linewidth]{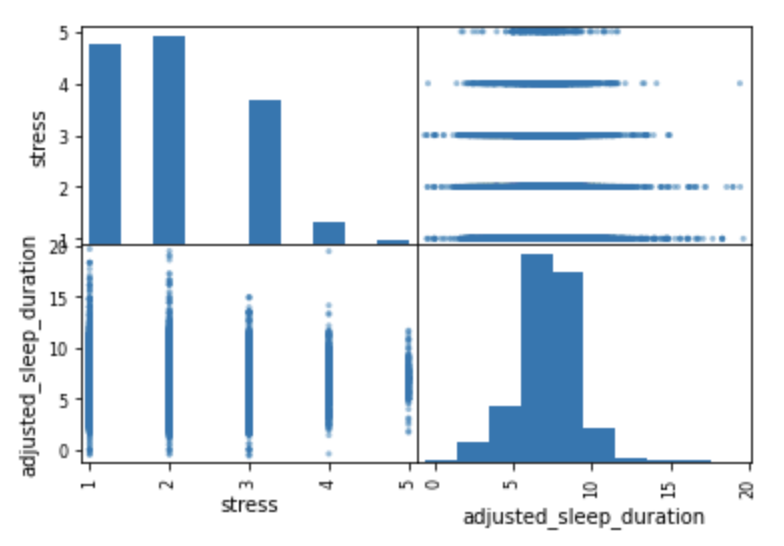}  
  \caption{Sleep duration vs Stress}
  \label{fig:sub-third}
\end{subfigure}
\begin{subfigure}{.5\textwidth}
  \centering
  % include fourth image
  \includegraphics[width=.8\linewidth]{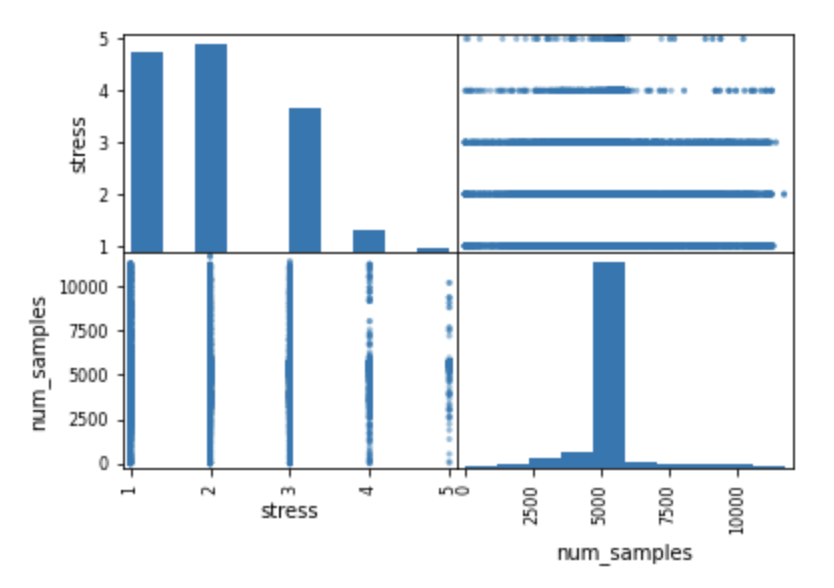}  
  \caption{Heart rate samples number vs Stress }
  \label{fig:sub-fourth}
\end{subfigure}
\caption{Relation between some of the features with the stress for imbalanced data}
\label{fig:ComparisonWithoutPersonalitySMOTE}
\end{figure}

    %\newpage
    \item \textbf{Classification}\\
    In Table \ref{tab:ClassificationWithoutPersonalityImbalanced}, we present the classification performance score details for \textit{without personality dataset} in original form. 
    We observed identical performance scores presented as in Table \ref{tab:ClassificationWithoutPersonalityImbalanced} when using $107$ out of $184$ features. Thus, to avoid the curse of dimensionality, we removed the ones that did not contribute to the overall accuracy. To better explain the results, we shared the first $20$ important ones in Figure \ref{featuresWithClassImbalance}.

     We applied RF classification to these selected $107$ features (biomarkers). The detailed classification results are in Table \ref{tab:ClassificationWithoutPersonalityImbalanced}. As we have a class imbalance, we obtained very low f1-scores for the $4th$ and $5th$ classes, which present the most highly stressful classes. However, for these classes, we have high precision and low recall. It can be interpreted as correctly predicted labels; however, actual instances belonging to these classes are misclassified. We observed lower f1-score even in classes with high instances ($1st$, $2nd$); $0.68$ and $0.60$ respectively. In addition, we provide a  confusion matrix in Figure \ref{fig:ConfusionWithoutPersonalityImbalanced}. Class $1$ is mostly confused with the $2nd$ class. In the confusion matrix, the percentages are distributed over each pair. Thus, the sum of all percentages corresponds to $100\%$.

        \begin{table}[!htb]
        \centering
        \caption{Without personality with class imbalance}
        \label{tab:ClassificationWithoutPersonalityImbalanced}
\begin{tabular}{rrrrr}
                      & \multicolumn{1}{l}{\textbf{precision}} & \multicolumn{1}{l}{\textbf{recall}} & \multicolumn{1}{l}{\textbf{f1-score}} & \multicolumn{1}{l}{\textbf{support}} \\
\multicolumn{1}{l}{}  & \multicolumn{1}{l}{}                   & \multicolumn{1}{l}{}                & \multicolumn{1}{l}{}                  & \multicolumn{1}{l}{}                 \\
\textbf{1}            & 0.65                                   & 0.71                                & 0.68                                  & 2244                                 \\
\textbf{2}            & 0.52                                   & 0.72                                & 0.60                                  & 2205                                 \\
\textbf{3}            & 0.70                                   & 0.35                                & 0.46                                  & 1611                                 \\
\textbf{4}            & 0.97                                   & 0.12                                & 0.21                                  & 251                                  \\
\textbf{5}            & 0.92                                   & 0.25                                & 0.39                                  & 44                                   \\
\multicolumn{1}{l}{}  &                                        &                                     &                                       &                                      \\
\textbf{accuracy}     &                                        &                                     & 0.60                                  & 6355                                 \\
\textbf{macro avg}    & 0.75                                   & 0.43                                & 0.47                                  & 6355                                 \\
\textbf{weighted avg} & 0.63                                   & 0.60                                & 0.58                                  & 6355                                
\end{tabular}
\end{table}

    \begin{figure}[!htb]
        \centerline{\includegraphics[width=0.65\columnwidth]{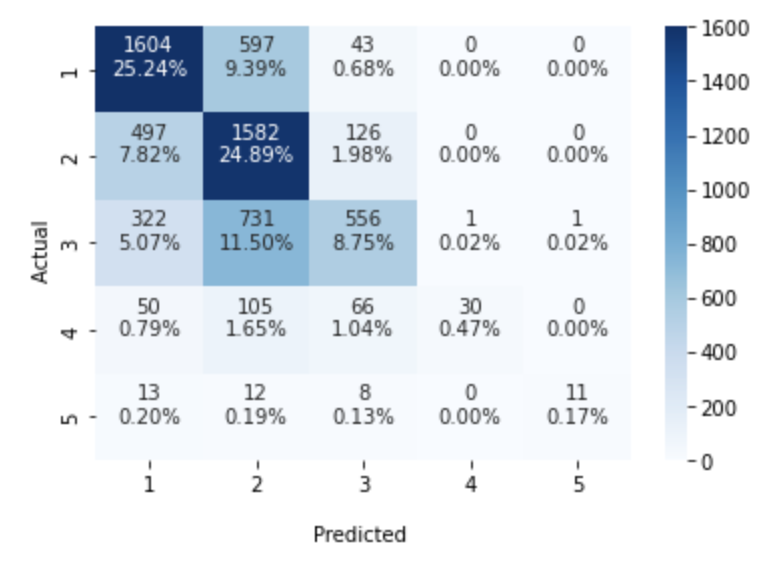}}
        \caption{Confusion matrix without personality with imbalance}
        \label{fig:ConfusionWithoutPersonalityImbalanced}
        \end{figure}

\end{itemize}

\newpage
\subsubsection{Performance after applying SMOTE}
\label{subSec:WithoutPersonalityAfterSMOTE}
To resolve the class imbalance, we applied the SMOTE technique. The idea is to produce synthetic data by keeping the same distribution of each class. Before applying SMOTE, the number of instances for each stress class was $10991$, $11430$, $7935$, $1197$, and $219$ for classes $1$ to $5$ respectively; after SMOTE, we have $11430$ instances for each class. Thus, the number of rows increases to $57150$ from $31772$.

\begin{itemize}
    \item \textbf{Most important biomarkers}\\
    Similarly, we applied RF feature selection and extracted most important $20$ biomarkers in Figure \ref{fig:featuresWithoutPersonalitySMOTE}. Even though the features selected in the imbalanced version are different, as we saw in Figure \ref{fig:sub-secondWithoutPersonality} that the ranking of the modalities is the same for the first ones, which are phone activity and sleep modalities. However, there is a slight change in the ranking of HR, daily, stress, and activity modalities.

     \begin{figure}[!htbp]
        \centerline{\includegraphics[width=\columnwidth]{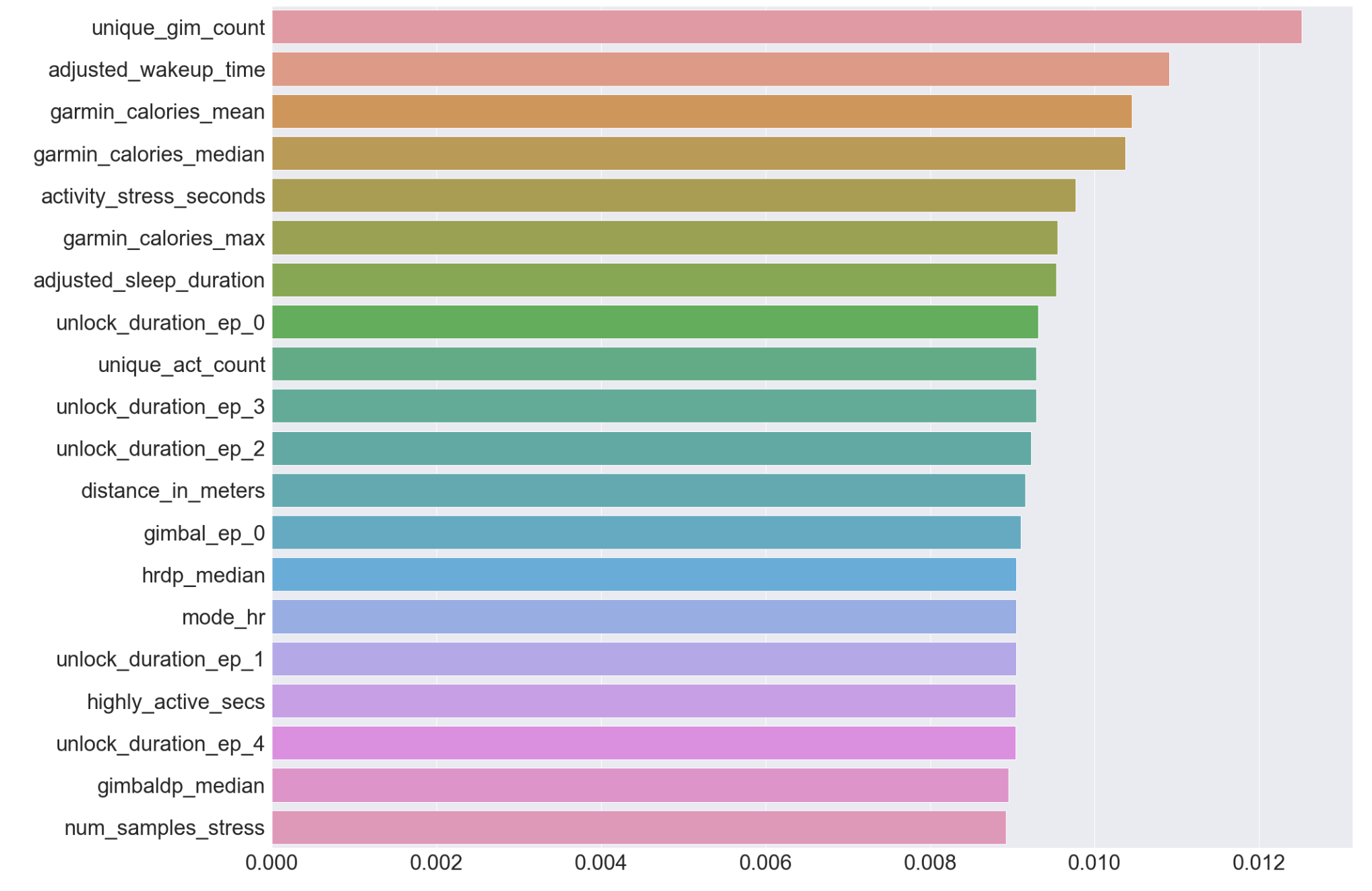}}
        \caption{Feature ranking after SMOTE without personality}
        \label{fig:featuresWithoutPersonalitySMOTE}
        \end{figure}
        
        \newpage
   \item \textbf{Classification}\\
     Classification result and confusion matrix can be seen in Table \ref{tab:ClassificationWithoutPersonalitySMOTE} and Figure \ref{fig:ConfusionWithoutPersonalitySMOTE}. We see the effect of balancing the classes. Overall class accuracy performance increased from $0.60$ to $0.78$. However, when we examine it in detail, we notice that this improvement comes from the $4th$ and $5th$ classes which have lower instances than other classes. The performance of the remaining classes was approximately the same except for the $3rd$ class, which has slightly increased.
    
    \begin{table}[!htb]
    \centering
    \caption{Without personality after SMOTE}
    \label{tab:ClassificationWithoutPersonalitySMOTE}
\begin{tabular}{rrrrr}
                      & \multicolumn{1}{l}{\textbf{precision}} & \multicolumn{1}{l}{\textbf{recall}} & \multicolumn{1}{l}{\textbf{f1-score}} & \multicolumn{1}{l}{\textbf{support}} \\
\multicolumn{1}{l}{}  & \multicolumn{1}{l}{}                   & \multicolumn{1}{l}{}                & \multicolumn{1}{l}{}                  & \multicolumn{1}{l}{}                 \\
\textbf{1}            & 0.66                                   & 0.70                                & 0.68                                  & 2306                                 \\
\textbf{2}            & 0.58                                   & 0.61                                & 0.59                                  & 2212                                 \\
\textbf{3}            & 0.75                                   & 0.62                                & 0.68                                  & 2305                                 \\
\textbf{4}            & 0.92                                   & 0.98                                & 0.95                                  & 2320                                 \\
\textbf{5}            & 1.00                                   & 1.00                                & 1.00                                  & 2287                                 \\
\multicolumn{1}{l}{}  &                                        &                                     &                                       &                                      \\
\textbf{accuracy}     &                                        &                                     & 0.78                                  & 11430                                \\
\textbf{macro avg}    & 0.78                                   & 0.78                                & 0.78                                  & 11430                                \\
\textbf{weighted avg} & 0.78                                   & 0.78                                & 0.78                                  & 11430                               
\end{tabular}
\end{table}
    
      \begin{figure}[!htbp]
        \centerline{\includegraphics[width=0.65\columnwidth]{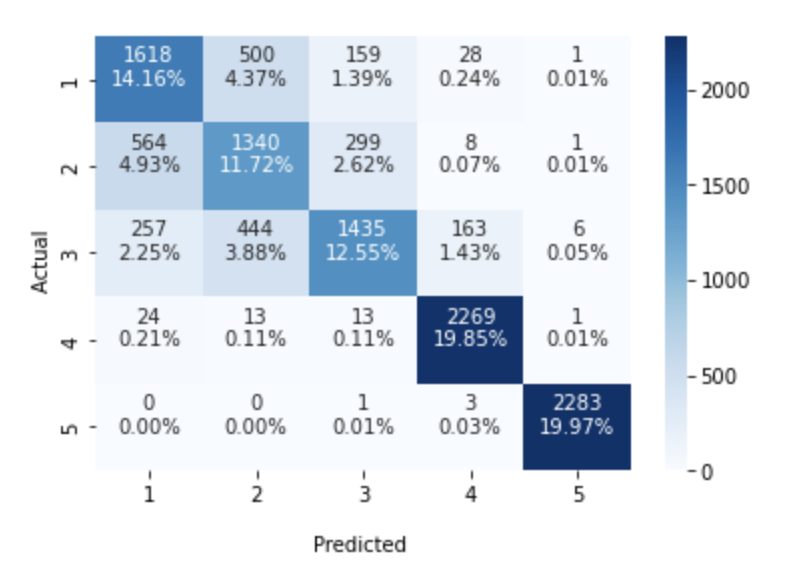}}
        \caption{Confusion matrix without personality balanced data}
        \label{fig:ConfusionWithoutPersonalitySMOTE}
        \end{figure}

\end{itemize}

        %\newpage
\subsection{With Personality dataset results}
In this experiment, we add personality and other parameters collected during the ground truth part into our parameter space. We repeat the same process. Here, we deal with $194$ independent variables, and our target is again the stress variable. 

\subsubsection{With Class Imbalance}
\begin{itemize}
    \item \textbf{Biomarkers}\\
Since we added new parameters after feature selection, we observed them in the new feature ranking in Figure \ref{fig:featuresPersonalityImbalanced}. We observed $7$ over $10$ recently added parameters among the most important biomarkers. The two overwhelming important biomarkers are anxiety and negative affection over stress. This result is coherent with the literature \cite{a18,a19}. It is stated that there are two types of stress; acute and chronic \cite{a18}. The triggering factor for the acute stress may be the anxiety \cite{a19}. Thus, it is expected that it is found among the most affecting factors.

The general ranking of the modalities is shown in Figure \ref{fig:ImlbalancedWithPersonality}. Therefore, we can say that sleep and phone activity modalities keep their places of importance, and recently added features occur in our ranking.
In Figure \ref{fig:ComparisonWithPersonalitySMOTE}, we share the \textit{anxiety, agreeableness, consciousness, neuroticism} versus stress distribution. There are slightly visible effects between stress classes, especially for the distinction of the $5th$ class from the rest. For instance, we can see an effect where the range of these features become narrow when stress level increase. We see the effect of the newly added parameters on the stress.

        \begin{figure}[!htbp]
        \centerline{\includegraphics[width=\columnwidth]{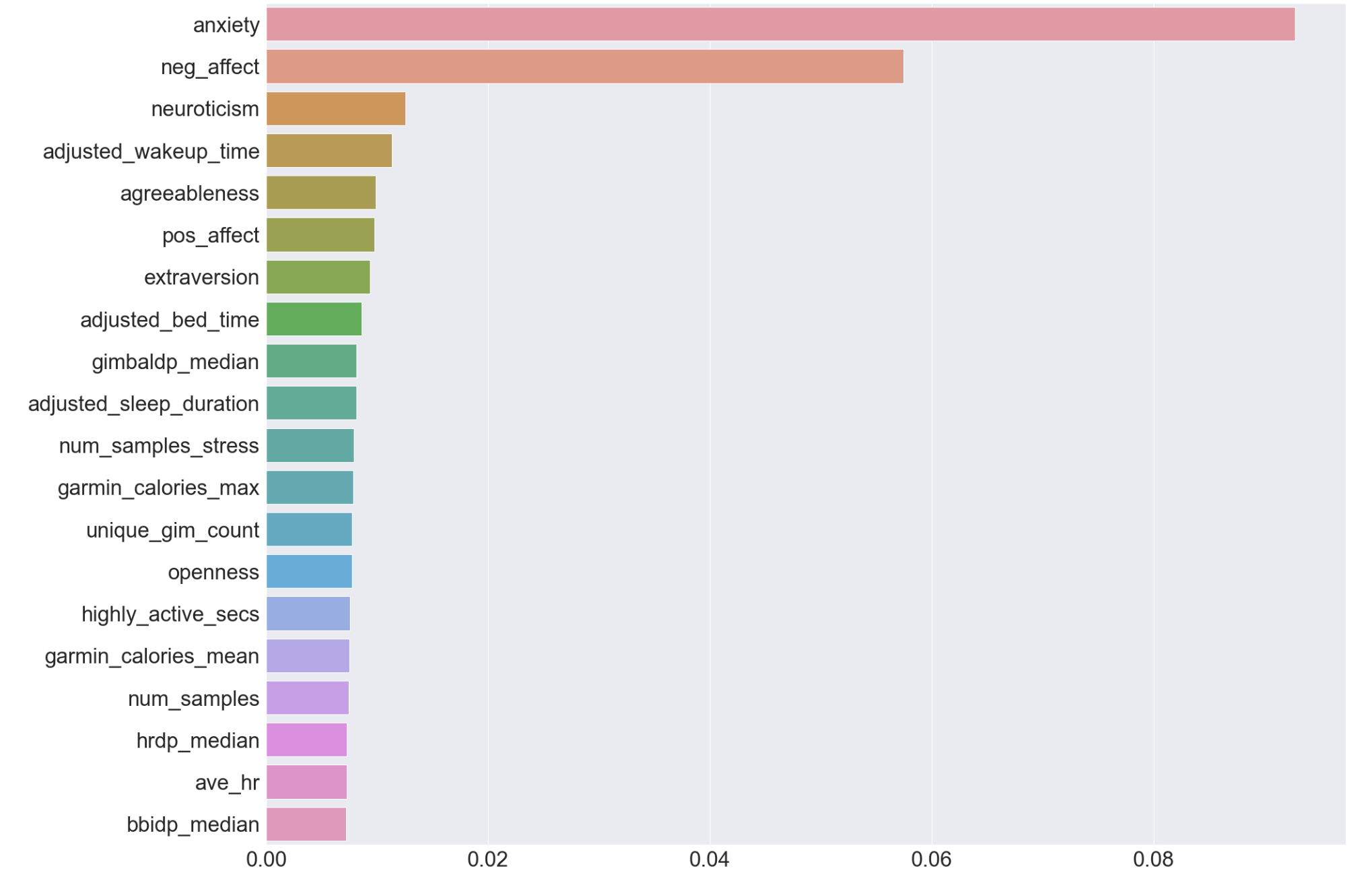}}
        \caption{Feature ranking with class imbalance with personality}
        \label{fig:featuresPersonalityImbalanced}
        \end{figure}

        \begin{figure}[!htb]
\begin{subfigure}{.5\textwidth}
  \centering
  % include first image
  \includegraphics[width=\linewidth]{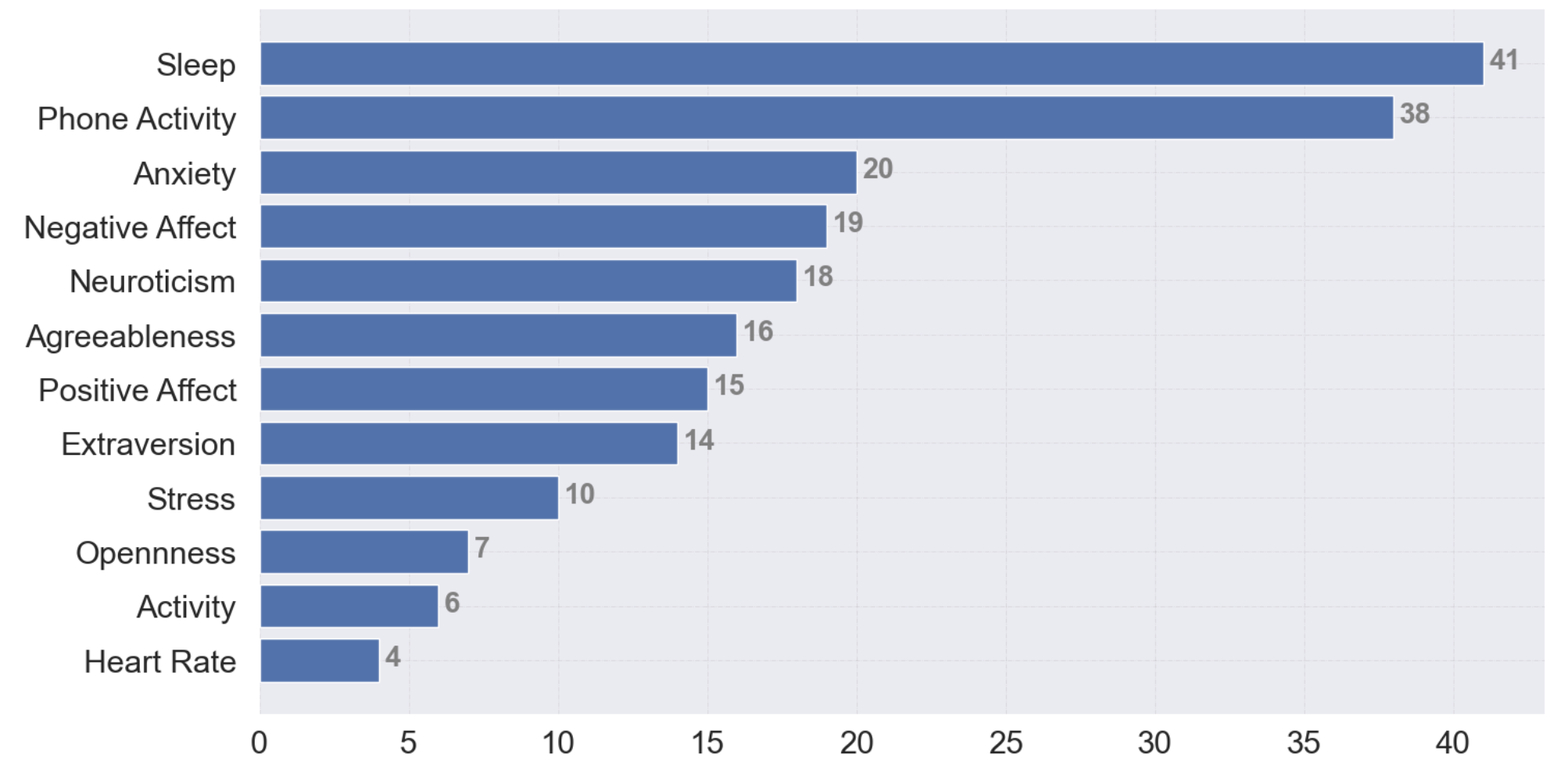}  
  \caption{With Class Imbalance}
  \label{fig:ImlbalancedWithPersonality}
\end{subfigure}
\begin{subfigure}{.5\textwidth}
  \centering
  % include second image
  \includegraphics[width=\linewidth]{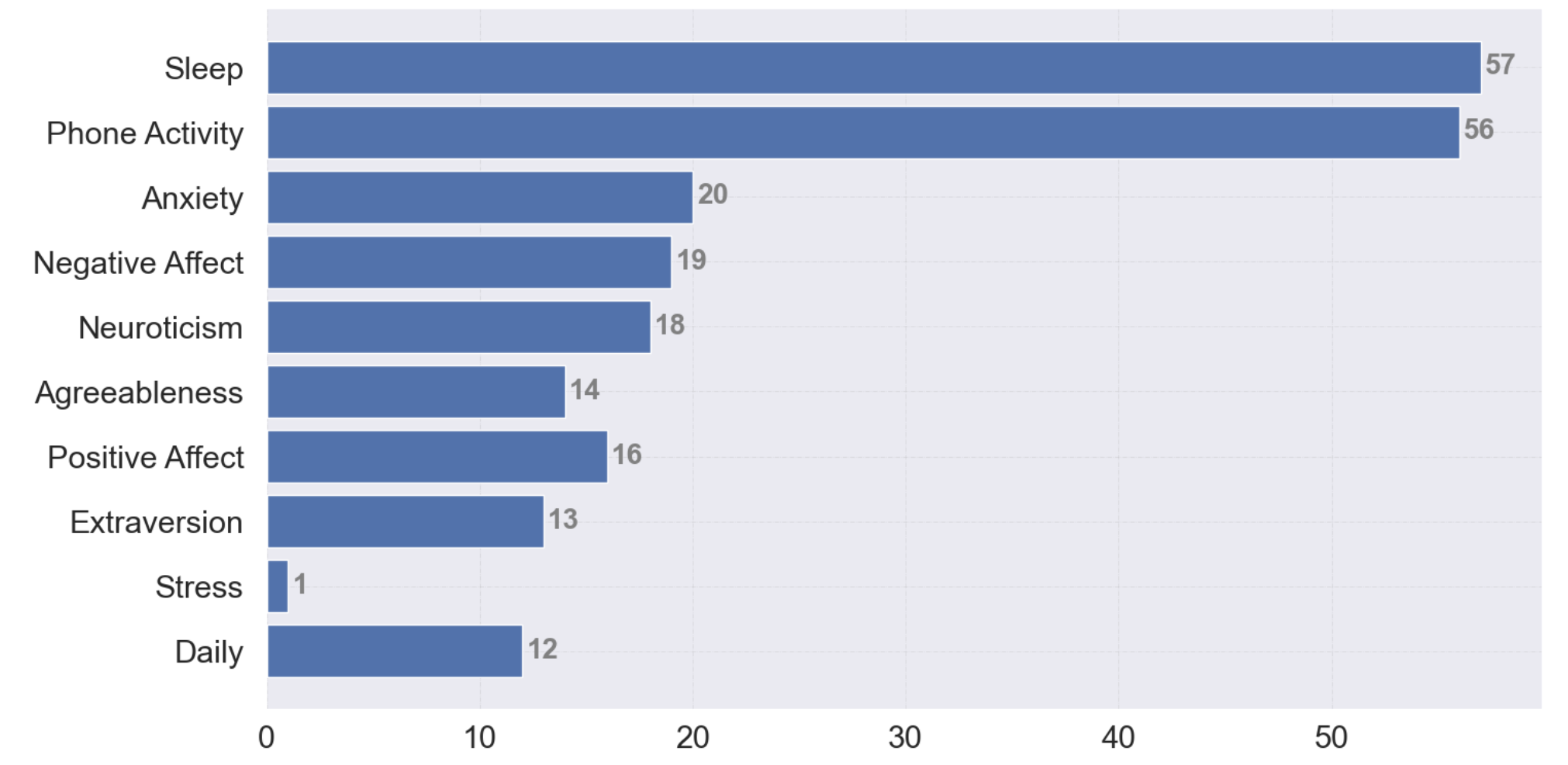}  
  \caption{After SMOTE application}
  \label{fig:SMOTEWithPersonality}
\end{subfigure}
\caption{Ranking of biomarkers' corresponding modalities: Personality Dataset}
\label{fig:RankingWithPersonality}
\end{figure}

\begin{figure}
\begin{subfigure}{.5\textwidth}
  \centering
  % include first image
  \includegraphics[width=.8\linewidth]{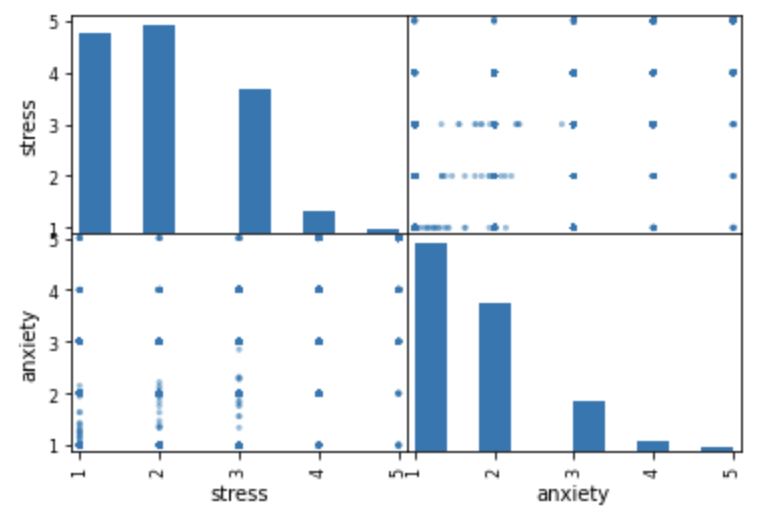}  
  \caption{Anxiety vs Stress}
  \label{fig:WakeupStress}
\end{subfigure}
\begin{subfigure}{.5\textwidth}
  \centering
  % include second image
  \includegraphics[width=.8\linewidth]{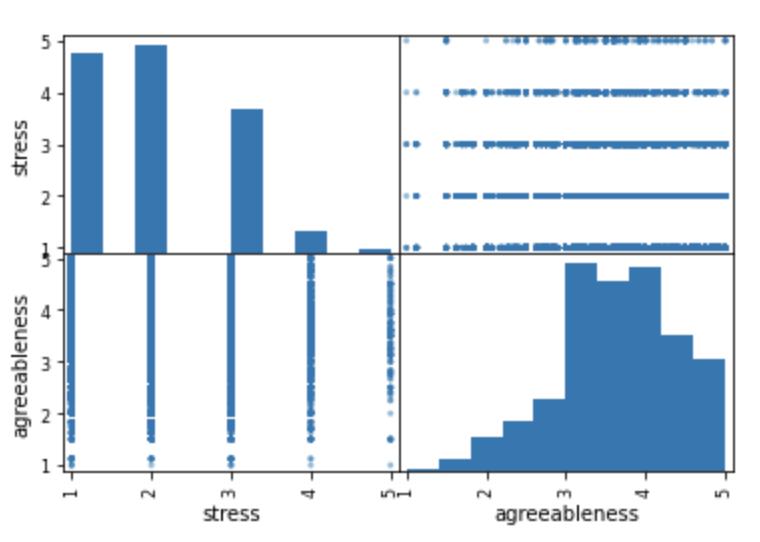}  
  \caption{Agreeableness vs Stress}
  \label{fig:sub-second}
\end{subfigure}

\begin{subfigure}{.5\textwidth}
  \centering
  % include third image
  \includegraphics[width=.8\linewidth]{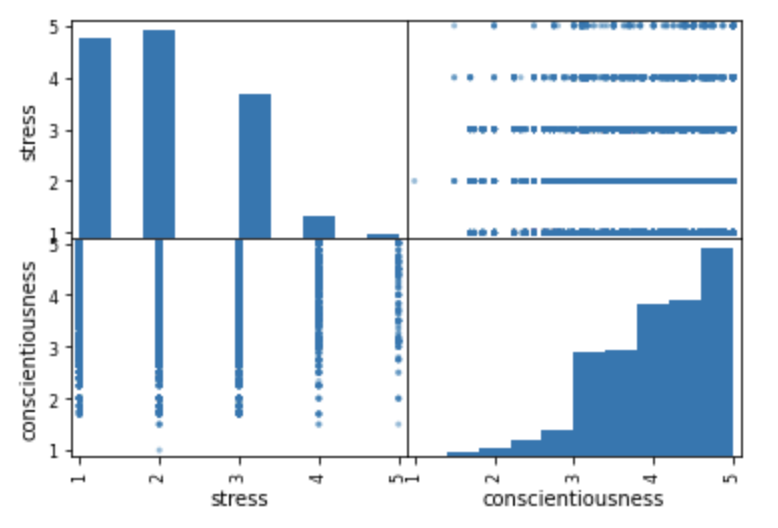}  
  \caption{Consciousnesses vs Stress}
  \label{fig:sub-third}
\end{subfigure}
\begin{subfigure}{.5\textwidth}
  \centering
  % include fourth image
  \includegraphics[width=.8\linewidth]{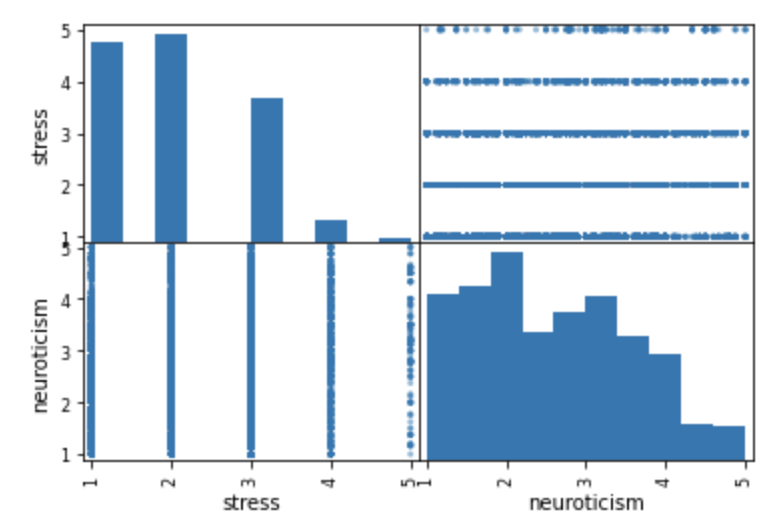}  
  \caption{Neuroticism vs Stress }
  \label{fig:sub-fourth}
\end{subfigure}
\caption{Pair relation some of the features with the stress for imbalanced data }%(With Personality Dataset)
\label{fig:ComparisonWithPersonalitySMOTE}
\end{figure}

  %\newpage 
    \item \textbf{Classification}\\
    In Table \ref{tab:performancePersonalityWithoutSMOTE}, we observe that even with a class imbalance on the target variable, we got higher f1 scores compared to the \textit{without personality} dataset. Thus, we can say that these parameters improve the overall result accuracy from $0.60$ to $0.71$. The detailed confusion matrix can be seen in Figure \ref{fig:COnfusionPersonalityWithoutSMOTE}.

    \begin{table}[!htb]
    \centering
     %\addtolength{\tabcolsep}{-7pt}
    \caption{With personality with class imbalance}
    \label{tab:performancePersonalityWithoutSMOTE}
    \begin{tabular}{rrrrr}
                      & \multicolumn{1}{l}{\textbf{precision}} & \multicolumn{1}{l}{\textbf{recall}} & \multicolumn{1}{l}{\textbf{f1-score}} & \multicolumn{1}{l}{\textbf{support}} \\
\multicolumn{1}{l}{}  & \multicolumn{1}{l}{}                   & \multicolumn{1}{l}{}                & \multicolumn{1}{l}{}                  & \multicolumn{1}{l}{}                 \\
\textbf{1}            & 0.76                                   & 0.86                                & 0.81                                  & 2244                                 \\
\textbf{2}            & 0.67                                   & 0.67                                & 0.67                                  & 2205                                 \\
\textbf{3}            & 0.69                                   & 0.67                                & 0.68                                  & 1611                                 \\
\textbf{4}            & 0.79                                   & 0.14                                & 0.23                                  & 251                                  \\
\textbf{5}            & 0.92                                   & 0.27                                & 0.42                                  & 44                                   \\
\multicolumn{1}{l}{}  &                                        &                                     &                                       &                                      \\
\textbf{accuracy}     &                                        &                                     & 0.71                                  & 6355                                 \\
\textbf{macro avg}    & 0.77                                   & 0.52                                & 0.56                                  & 6355                                 \\
\textbf{weighted avg} & 0.71                                   & 0.71                                & 0.70                                  & 6355                                
\end{tabular}
\end{table}

        \begin{figure}[!htbp]
        \centerline{\includegraphics[width=0.65\columnwidth]{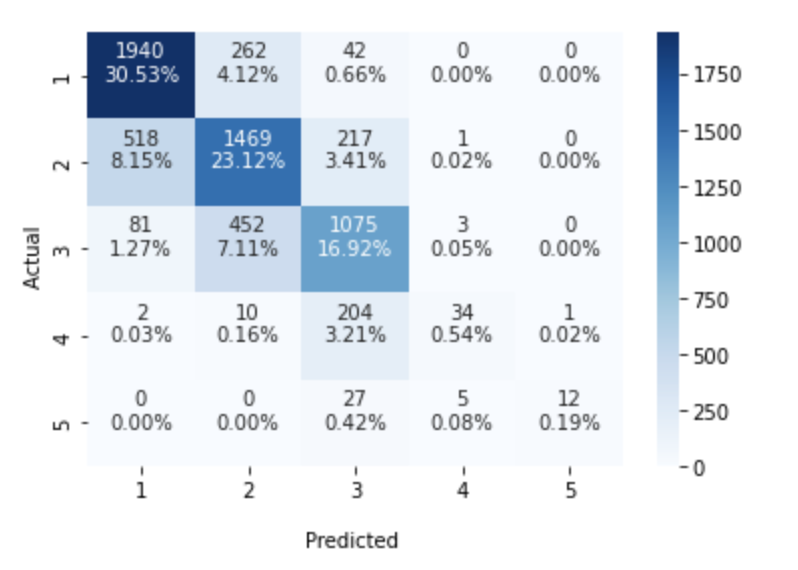}}
        \caption{Confusion matrix with personality with imbalance}
        \label{fig:COnfusionPersonalityWithoutSMOTE}
        \end{figure}      
    
\end{itemize}

%\newpage        
\subsubsection{Performance after applying SMOTE}
\begin{itemize}
    \item \textbf{Biomarkers}\\
   Again, we apply SMOTE to solve the imbalance issue in stress classes. The parameter space is the only difference from the \textit{without personality} dataset. The row numbers after the SMOTE remain the same compared to \textit{without personality} dataset. Extracted biomarkers can be seen in Figure \ref{featuresSMOTE_personality}. After SMOTE, the importance of order changed. The details are in Figure \ref{fig:SMOTEWithPersonality}. Phone-related features rank up. Sleep features are important with different ratios.

     \begin{figure}[!htbp]
        \centerline{\includegraphics[width=\columnwidth]{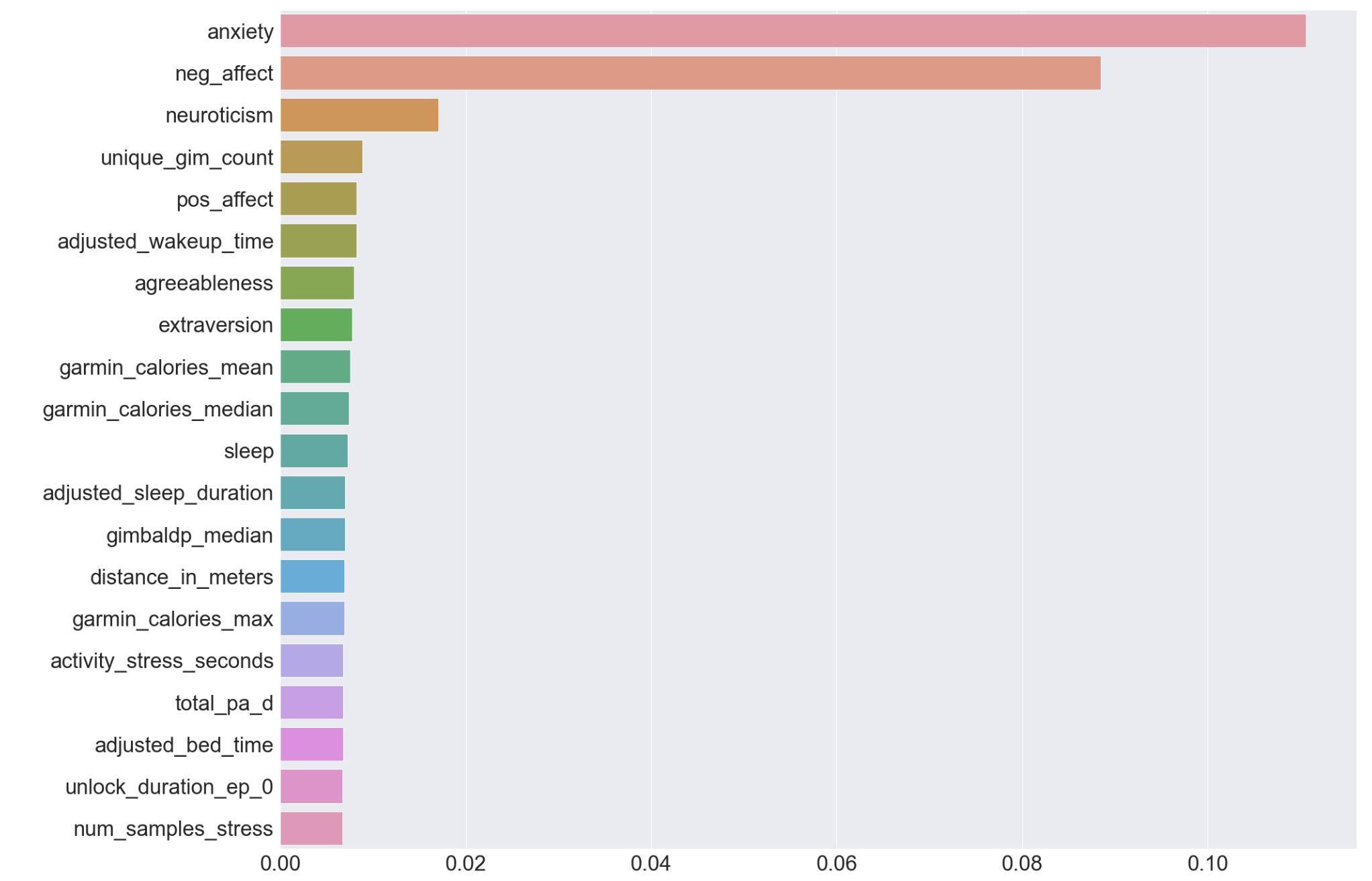}}
        \caption{Feature ranking after SMOTE with personality}
        \label{featuresSMOTE_personality}
        \end{figure}

   \item \textbf{Classification}\\
    We see the effect of balanced classes. Overall performance in accuracy is increased from $0.71$ to $0.85$. However, when we examine in detail, we notice that this improvement comes from the $4th$ and $5th$ classes, which have lower instances than other classes similar to \textit{without personality dataset results} in Section \ref{subSec:WithoutPersonalityAfterSMOTE} after SMOTE application. The other classes' performance remained quite the same except for the $3rd$ one, which increased from $0.68$ to $0.81$ f1-score. Performance details can be seen in Table \ref{tab:PerformancePersonalityMSOTE} and confusion matrix in Figure \ref{fig:ConfusionPersonalitySMOTE}.

     \begin{table}[!htb]
    \centering
    \caption{With personality after SMOTE}
    \label{tab:PerformancePersonalityMSOTE}
\begin{tabular}{rrrrr}
                      & \multicolumn{1}{l}{\textbf{precision}} & \multicolumn{1}{l}{\textbf{recall}} & \multicolumn{1}{l}{\textbf{f1-score}} & \multicolumn{1}{l}{\textbf{support}} \\
\multicolumn{1}{l}{}  & \multicolumn{1}{l}{}                   & \multicolumn{1}{l}{}                & \multicolumn{1}{l}{}                  & \multicolumn{1}{l}{}                 \\
\textbf{1}            & 0.77                                   & 0.87                                & 0.82                                  & 2306                                 \\
\textbf{2}            & 0.71                                   & 0.62                                & 0.66                                  & 2212                                 \\
\textbf{3}            & 0.82                                   & 0.79                                & 0.81                                  & 2305                                 \\
\textbf{4}            & 0.95                                   & 0.99                                & 0.97                                  & 2320                                 \\
\textbf{5}            & 1.00                                   & 1.00                                & 1.00                                  & 2287                                 \\
\multicolumn{1}{l}{}  &                                        &                                     &                                       &                                      \\
\textbf{accuracy}     &                                        &                                     & 0.85                                  & 11430                                \\
\textbf{macro avg}    & 0.85                                   & 0.85                                & 0.85                                  & 11430                                \\
\textbf{weighted avg} & 0.85                                   & 0.85                                & 0.85                                  & 11430                               
\end{tabular}
\end{table}

    \begin{figure}[!htbp]
        \centerline{\includegraphics[width=0.65\columnwidth]{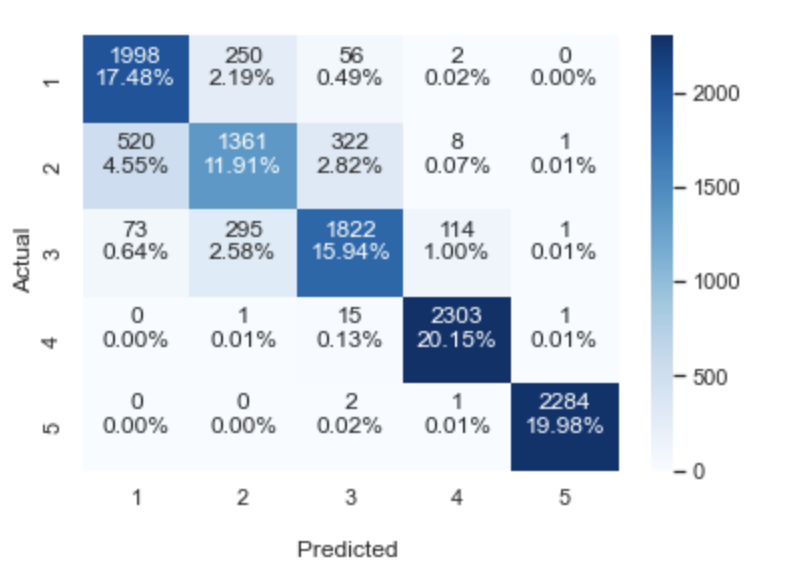}}
        \caption{Confusion matrix with personality balanced data}
        \label{fig:ConfusionPersonalitySMOTE}
        \end{figure}
    
\end{itemize}

\subsection{Discussion and Comparison}
\label{sec:discuss}
Considering our two different experiments and their sub-scenarios, we obtained higher classification results by including ground truth parameters about personality in our parameter space. These parameters are extraversion, openness, neuroticism, agreeableness, consciousness, and mood-related parameters, such as positive and negative affect. In addition, we observe that they are selected among the most important parameters after feature ranking impacting the stress state and increasing the classification scores. For instance, we obtain $0.60$ accuracy (Table \ref{tab:ClassificationWithoutPersonalityImbalanced}) when they are not included the analysis, while we have $0.71$ accuracy (Table \ref{tab:performancePersonalityWithoutSMOTE}) when we add them. The classification performances are even higher when we apply SMOTE and solve the class imbalance problem, which is $0.78$ (Table \ref{tab:ClassificationWithoutPersonalitySMOTE}) when these parameters are not considered, and $0.85$ accuracy is achieved (Table \ref{tab:PerformancePersonalityMSOTE}) when they are taken into account. The most important ones are anxiety and negative affection. Intuitively, these are highly related to stress levels. Therefore, their usage for further analysis may be a helpful approach. Besides the personality-related parameters, sleep and phone activity-related markers are the most important ones in all cases (Figure  \ref{fig:RankingBiomarkerModalitiesWithoutPersonality} and Figure \ref{fig:RankingWithPersonality}). In \cite{a20}, authors found a strong relationship between sleep and stress. Also, in \cite{a21},
health-related factors such as sleep patterns are found among the important ones affecting stress. Moreover, in study \cite{a22}, researchers reveal the relationship between positive affect and well-being with sleep. Thus, they are also related to stress due to the relationship between sleep and stress. In \cite{a23}, again, sleep is found to be related to the mood state. Furthermore, the relationship between exercise and anxiety decrease is examined and found related. Thus, our results aligned with the literature.

We also observed significant performance improvement after applying SMOTE since the number of instances per stress class became similar. We can better recognize the classes with lower instances. When we look in detail, we observe that the f1-score from the classes of lower instances is lower than the other classes. This is due to the class imbalance problem. Since we have fewer instances than other classes, the model fails to recognize them well. After we apply SMOTE to solve the class imbalance problem by generating data similar to the current measurements, we obtain higher final class f1 scores. As it was not the case for the classes with higher instances, they had lower precision rates than the others, and their final class scores did not change much after the SMOTE application.

Comparing classifier performance with other stress level classification studies in the literature is also important. In \cite{a17}, the authors prepared a detailed survey on stress detection in daily life scenarios using wearables. They present the performance in the literature along with the devices, sensors, and methods used. We observe that we achieve similar and even better results considering the ambulatory scenarios in the literature \cite{a17} even with a higher number of classes. In \cite{a24},  they used speech data, applied GMM (Gaussian Mixture Model), and obtained $80.5\%$ accuracy in the stress and relaxed states classification. Another high result is presented in \cite{a25}. They received $76\%$ accuracy using \textit{blood volume pulse, skin temperature, electro-dermal activity (EDA), heart rate variability, heart rate} parameters applying Random Forest on two classes (stressed, relaxed) problem.  

In addition to RF, we also applied deep learning (DL) methods as they performed better in literature \cite{b29}. We used MLP (Multi-layer Perceptron). We performed hyper-parameter optimization with grid search in scikit-learn and found that activation \textit{relu}, alpha \textit{$0.0001$}, hidden layer \textit{$30$}, solver \textit{adam} reveal the best performance. However, overall class performances for our scenarios with MLP compared to RF are lower by about $20\%$. For instance, for our best case (with personality and with SMOTE), we have $85\%$ with RF but $63\%$ with MLP. As these results were not promising, we did not include their details in this study.

%\newpage
\section{Conclusion and Future Work}
\label{sec:Conclusion}
In this study, we examined the digital biomarkers related to stress state since they become valuable information for well-being. We used the Tesserae dataset, collected from office workers. There have been several different modalities. We used the Machine Learning technique to examine these markers, specifically the Random Forest algorithm. We extracted them by dealing with multi-modal data. Also, we made classification with selected biomarkers which refer to high importance in \textit{stress} class. We applied SMOTE technique to solve the imbalance issue in the target class. We also analyzed the impact of using personality-related parameters. In all cases, phone activity and sleep modalities-related markers are the most relevant to \textit{stress}. Highest classification accuracy is $85\%$ using \textit{with personality dataset} after applying SMOTE. Our findings confirm the literature that examines the statistical relationship between sleep, well-being, stress, positive affect, mood, and physical activity modalities \cite{a20,a21,a22,a23}. We contribute by working all these modalities in one study in a multi-modal manner by applying machine learning methods over wearable and survey data collected in an unrestricted environment.

We want to point out some improvements for further studies. We excluded collected wearable and mobile phone data without ground truth. This amount is vast since there is almost one-year collection of data from wearables for some participants. Thus, one further approach may be an imputation of ground truth data corresponding to these data to use in the processing. In addition, we did not consider the time-related aspects in this study. For instance, there may be changes in stress levels according to the days of the week. It is intuitively expected that one may have a lower stress level during weekends and holidays. Furthermore, we will explore different DL methods, such as CNN and LSTM. We will consider these aspects in our future studies.

\section*{Acknowledgment}

Tübitak Bideb 2211-A academic reward is gratefully acknowledged. This work is supported by the Turkish Directorate of Strategy and Budget under the TAM Project number 2007K12-873.

\newpage

\end{document}